\documentclass[preprint,12pt]{article}

\makeatletter

\usepackage[utf8]{inputenc} 
\usepackage[T1]{fontenc}    
\usepackage[hidelinks]{hyperref}       
\usepackage{url}            
\usepackage{booktabs}       
\usepackage{amsfonts}       
\usepackage{nicefrac}       
\usepackage{microtype}      
\usepackage{xspace}
\usepackage{enumerate}
\usepackage{xcolor}
\usepackage{graphicx}

\newcommand{\prob}{\mathbb{P}}  

\title{An interpretable alternative to the F1 measure}

\author{David J. T. Sumpter \\ Department of Information Technology, Uppsala University, Sweden}

\begin{document}

\maketitle

\begin{abstract}
Performance metrics are more useful if they can be explained in words. The area under the ROC curve (AUC) is interpretable, as the probability that a randomly chosen image containing a cat scores higher than a randomly chosen image without one, but it is pushed towards one by adding images without cats that are easy to classify. The F1-score avoids this problem, but cannot be interpreted in words in the same way. We propose an interpretable metric that answers the following question: when an algorithm is tuned to a threshold so that it can no longer distinguish the images it labels as cats from the images that really contain cats, how often does an image labelled as a cat actually contain one? We call this precision at the indistinguishability threshold (PIT), and show that, unlike AUC, it is unaffected by adding easy cases. We also put F1 into words: maximising F1 approximately maximises the probability that a randomly chosen real cat is labelled as a cat and a randomly chosen labelled image is a cat. This is one part of the condition that defines the indistinguishability threshold, which helps explain why F1 works well in practice. What it leaves out is how the algorithm ranks the images it wrongly labels as cats. The same construction can be used to tune the trade-off between false positives and false negatives in words. For example, if missing cats is more costly than false alarms, the threshold can be lowered until a real cat scores higher than an image labelled as a cat 60\% of the time, rather than half the time, and the resulting precision and recall reported in the same terms. On a benchmark data set of mammograms, two classifiers with the same AUC differ by 13 percentage points in PIT. We suggest reporting PIT wherever F1 is currently reported.
\end{abstract}

\section*{Introduction}

Imagine you have created a machine learning model that assigns a score, predicting whether an entity belongs to one of two classes (positive and negative). For example, the input might be a picture and the output whether or not the picture contains a cat. The scores are larger if your model evaluates the object as more likely to belong to the positive class (i.e if it labels the image as more likely to contain a cat). Such a scoring system explicitly underlies, for example, logistic regression, the output of a neural network and discriminant analysis \cite{smlbook}. Other, non-parametric machine learning methods, such as $k$-nearest neighbours \cite{fix1989discriminatory}, don't have an explicit score, but often have a parameter (e.g. $k$) which can be tuned in a way that mimics a threshold.

\begin{figure}[p]
\centering
\includegraphics[scale=1]{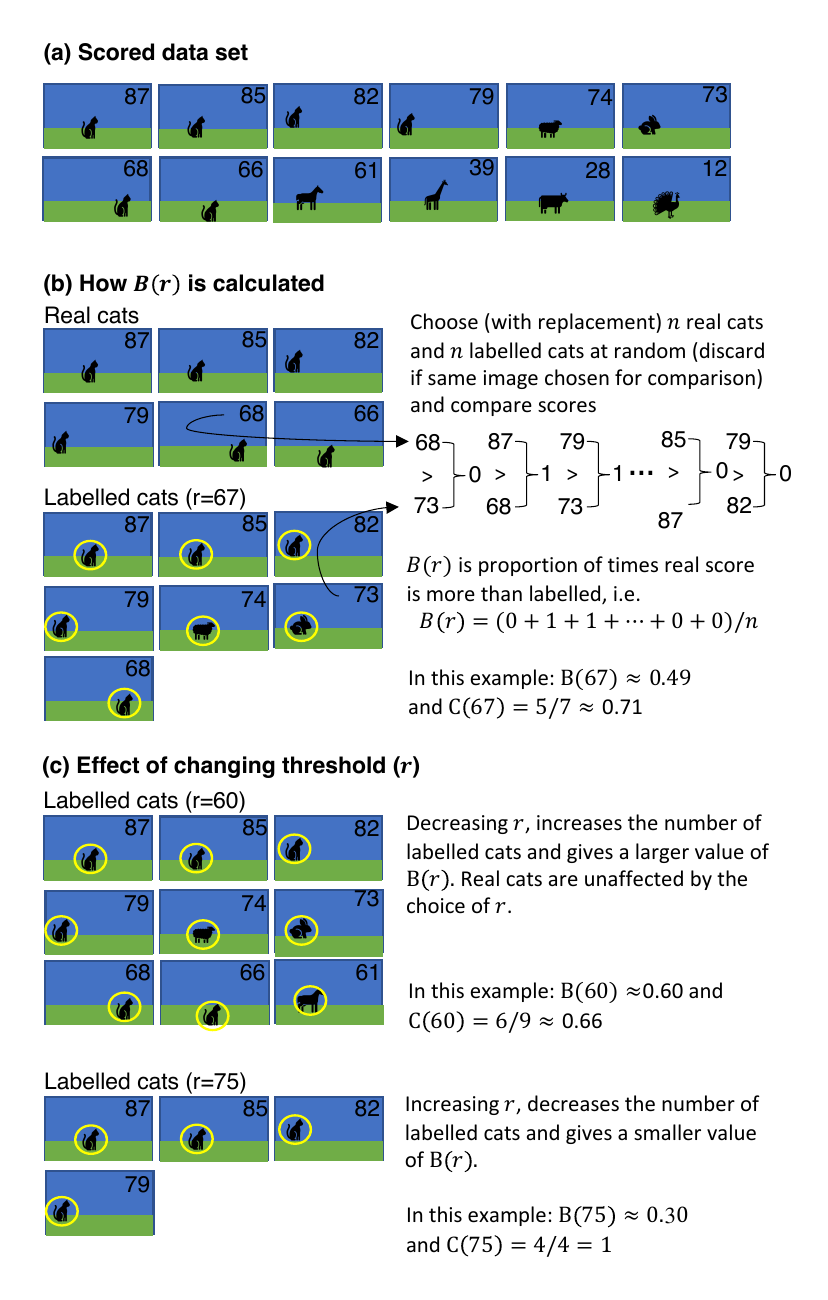}
\caption{Illustration of the method described in this paper} \label{fig:methodill}
\end{figure}

The method we propose for evaluating performance is as follows. We start, in figure \ref{fig:methodill}a, with a data set in which every image is assigned a score. We first investigate what happens when we set a threshold score, $r$, above which we label the image as containing a cat (positive), below which we label the image as not containing a cat (negative). For any given threshold we compare those images which we label as cats (given the threshold) with those containing real cats. Specifically, we choose (with replacement) $n$ real cats and $n$ labelled cats at random and compare scores (figure \ref{fig:methodill}b). In cases where we happen to pick the same two images, we pick two new images for that particular comparison. This gives a measure $B(r)$: the probability a real cat scores more than a labelled cat for threshold $r$.

We then identify a specific threshold score, $r_b$, such that if we were to choose one image that the algorithm classifies as containing a cat and a second image that really does contain a cat, then the probability that the score of the image containing the cat is the largest of the two scores is 50\%. We define this as $r_b$, where  $B(r_b) \approx 0.5$. In the example, in figure \ref{fig:methodill}b $r_b=67$. A smaller threshold gives $B(r) > 0.5$ and a larger threshold gives $B(r) < 0.5$ (figure \ref{fig:methodill}c). We call $r_b$ the {\it indistinguishability threshold}: because the set of images labelled as cats is indistinguishable from the set of images that really contain cats. When $r=r_b$, if we pick (at random) two images from our data set, one which is labelled as a cat and the other that is known to contain a cat, then the probability that the algorithm will assign the image with the cat the highest score is one half.

To measure the performance of your algorithm, simply calculate the proportion of images the algorithm classifies as containing a cat do actually contain a cat. We denote this measure $C(r_b)$ and call it {\it precision at the indistinguishability threshold}, or PIT.

There already exist a number of ways of evaluating classification algorithms. Wikipedia lists a somewhat bewildering array of 17 tests, with a total of 27 different names, for such problems.  The most commonly used method is the area under the Receiver-Operator Curve, commonly referred to as AUC \cite{obuchowski2018receiver, vanderlooy2008critical, majnik2013roc}. Much of its appeal lies in its simple interpretation: it is the probability that a randomly chosen member of the positive class scores higher than a randomly chosen member of the negative class \cite{bamber1975area, hanley1982meaning}. But AUC also has well-known limitations. It summarises performance over all thresholds, including those at which a classifier would never be used \cite{lobo2008auc}; it can favour one classifier over another even when the other performs better over almost all thresholds \cite{hand2009measuring}; and when the negative class is large it can give an overly optimistic view of how well a classifier identifies the positive class \cite{davis2006relationship, saito2015precision}. Responses to these limitations include the partial AUC, which restricts attention to a range of false positive rates \cite{mcclish1989analyzing, dodd2003partial, carrington2020new}, and precision-recall curves, which focus on how often positive labels are correct \cite{davis2006relationship, saito2015precision, flach2015precision}.

Another popular choice is the F1-score \cite{sasaki2007truth}, the harmonic mean of precision and recall. Because it ignores the members correctly labelled negative, it avoids the problem that AUC has with unbalanced data, in which many negatives are easy to classify \cite{smlbook}. But, unlike AUC, the F1-score has no simple interpretation in words, although it often works well in practice. Hand and Christen showed that it is a weighted arithmetic mean of precision and recall in which the weights depend on the classifier being evaluated, rather than on the problem \cite{hand2018note}, and Hand, Christen and Kirielle proposed a transformation of it, F*, with a more direct interpretation \cite{hand2021f}.

Measures such as precision and the F1-score are calculated at a particular threshold, so their values depend on how that threshold is chosen, even when the choice is left implicit, as when a default threshold is used or the F1-score is reported at its maximum. Common choices include the threshold which maximises the F1-score \cite{lipton2014optimal, koyejo2014consistent}; the threshold at which precision equals recall, known as the break-even point or R-precision \cite{manning2008introduction}; the threshold which maximises Youden's index, the difference between the true and false positive rates \cite{youden1950index}; and, when the costs of the two types of error are known, the threshold which minimises the expected cost \cite{elkan2001foundations}. In the quantification literature, which aims to estimate the number of members of each class, thresholds are also chosen by simple rules, such as setting the true positive rate to one half \cite{forman2008quantifying}. Of these, the break-even point is closest to the approach we take here. Since precision equals recall exactly when the number of members labelled positive equals the number of members of the positive class, the precision at the break-even point (PRBEP) can be stated in words: if the algorithm labels as many members positive as there are positives, what proportion of them are correct? Our threshold also matches the set labelled positive to the positive class, but by the ranking of their scores rather than by their size. Hern\'andez-Orallo, Flach and Ferri showed that many performance measures can be understood as the combination of a method for choosing a threshold and a measure of loss at that threshold \cite{hernandez2012unified}. 

In this paper, we propose a new method for choosing a threshold --- the one outlined above and presented in more detail below --- and argue that precision at this threshold, PIT, is an interpretable alternative to the F1-score. Like the F1-score, it is not inflated by cases that are easy to classify; like AUC, it can be explained in words. We first retell the interpretation of AUC, since the same argument, comparing randomly chosen pairs of members, underlies our measure. We then define the indistinguishability threshold and PIT, and test PIT on artificial and benchmark data. Finally, we use the same construction to put the F1-score into words. Maximising the F1-score approximately maximises one part of the condition that defines the indistinguishability threshold, which helps explain why the F1-score works as well as it does, and shows what it leaves out.

\section*{Results}

\subsection*{Area under the curve} 
\label{sec:AUC}

To get warmed up, we now show that the area under the ROC curve (AUC) measures the probability that a member of the positive class scores higher than a member of the negative class. This has been known since Bamber \cite{bamber1975area}, and was made widely known by Hanley and McNeil \cite{hanley1982meaning}. We retell it here since the argument, which compares the scores of randomly chosen pairs of members, is the same one we use to construct a new measure in the sections that follow.

To calculate this probability, we pick one member of the positive class at random and one member of the negative class at random and look to see which of them has the higher score. As a concrete example, imagine an algorithm is shown two images, one which contains a cat, another which does not. The task is to pick the image containing the cat. We call this probability 
\[
A = \prob(S_P>S_N)
\]
where $S_P$ is the score of a randomly chosen positive member and $S_N$ is the score of a randomly chosen negative member. The value of $A$ can be measured directly, simply by randomly picking pairs of members of negative and positive classes and testing which scores higher. The test outcome is then the proportion of cases in which the positive member receives a higher value than the negative member.

We can also measure the paired comparison test over all our scores. Let $F_P(r)$ be the cumulative distribution of scores of the positive class, i.e. 
\begin{equation}
F_P(r) = \prob(S_P < r) = \int_{s=-\infty}^{r} f_P(s) ds \label{eq:ScorefPr}
\end{equation}
where $f_P(r)$ is thus the probability density function of $S_P$. Likewise, we let $F_N(r)$ to be the cumulative distribution of scores the negative class, i.e. 
\begin{equation}
F_N(r) = \prob(S_N < r) = \int_{s=-\infty}^{r} f_N(s) ds \label{eq:ScorenPr}
\end{equation}
The value $A=\prob(S_P>S_N)$ can be written as the double integral
\begin{equation}
A = \int_{r=-\infty}^{\infty} \int_{s=r}^{\infty}  f_N(r) \cdot  f_P(s) ds dr  \label{eq:SPgreaterSN}
\end{equation}
which is equivalent to
\begin{equation}
A = \int_{r=-\infty}^{\infty}  f_N(r)  \int_{s=r}^{\infty}  f_P(s) ds dr  = \int_{r=-\infty}^{\infty}  f_N(r) \cdot  \left(1 -  F_P(r) \right) dr \label{eq:SPgreaterSN2}
\end{equation}
We can use this equation to calculate $A$ by numerical integration. 

We can also use this formulation to derive the equivalence between the $A$ test and AUC. To do so, note that the true positive rate for a given threshold $r$ is given by $v(r)=1-F_P(r)$ and that the false positive rate for a given threshold $r$, is given by $u(r)=1-F_N(r)$. Then we see that
\[
A  =  \int_{r=-\infty}^{\infty} v(r) f_N(r) dr  
\]
To change variable, write $x$ for a false positive rate and let
\[
\mathrm{ROC}(x) = v\left(u^{-1}(x)\right)
\]
be the true positive rate at the threshold at which the false positive rate is $x$, where $u^{-1}$ is the inverse of the function $u(r)$ (we assume that $u(r)$ is strictly decreasing, so that this inverse exists). The function $\mathrm{ROC}(x)$ is the ROC curve. Setting $x = u(r)$, so that $dx = -f_N(r)\, dr$ and $x$ decreases from one to zero as $r$ increases, gives
\begin{equation}
A  = \int_{x=0}^{1} \mathrm{ROC}(x)\, dx . \label{eq:AUCdef}
\end{equation}
Thus $A$ is also the area under the ROC curve.

\begin{figure}[tp]
\centering
\includegraphics[scale=1]{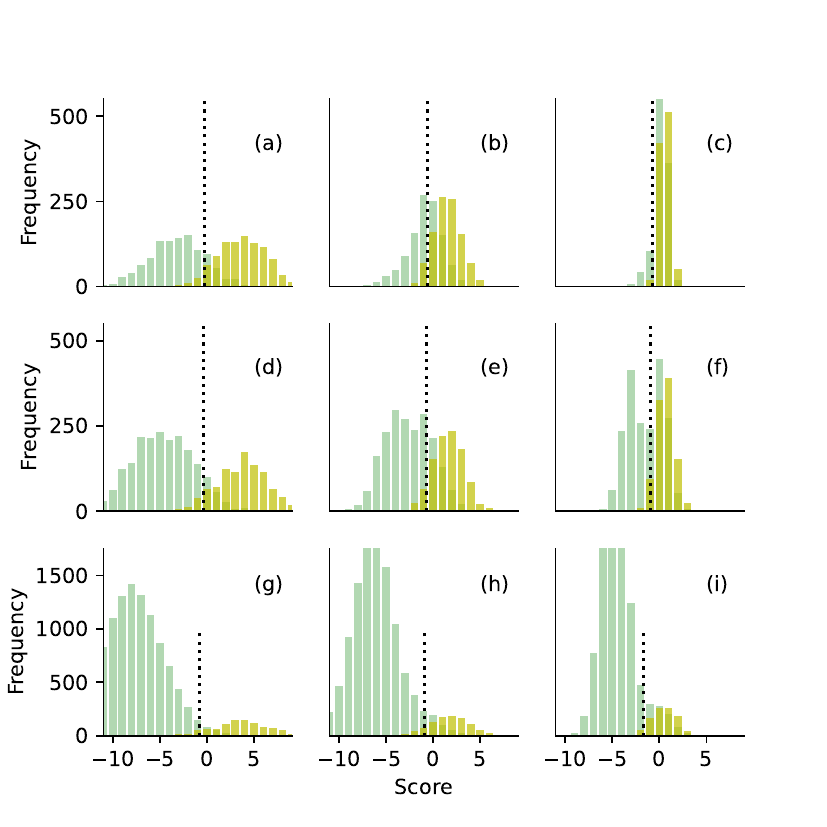}
\caption{The score distributions for the nine example data sets we look at here. Green are members of the negative class, yellow members of the positive class. The dotted line shows $r_b$, the solution $B(r)=1/2$ (see equation \ref{balanced})} \label{fig:Distribution}
\end{figure}

\subsection*{Artificial data sets}

To see how $A$, and the further measures introduced below, behave in practice, we use nine artificial data sets, shown in figure \ref{fig:Distribution}. The positive class ($Y_i=1$) is created by first generating 1000 input values, $X_i$, with a mean $m_p=10$ and standard deviation $\sigma_p=2$. Similarly, the negative class ($Y_i=-1$) consists of 1000 `difficult' data points with values, $X_i$, which have mean scores (a,d,g) $m_n=5$; (b,e,h) $m_n=7$; (c,f,i) $m_n=9$ and (in all cases) standard deviation $\sigma_n=2$, along with (a,b,c) 100; (d,e,f) 1000; (g,h,i) 10000 `easy' data points, $X_i$, with mean 2 and standard deviation 2.  We then performed logistic regression on $X_i$ to predict $Y_i$ and calculated the log-likelihood for every individual point. These log-likelihoods are taken to be the scores, $S_i$, output by the model and are plotted in figure \ref{fig:Distribution}. 

The choice of these data sets is to give a range of difficulty levels. The data in the left column correspond to relatively easy classification tasks. The overlap between the green (negative) and yellow (positive) is small. The data in the right column is much more difficult, with a large overlap in scores for negative and positive. The datasets at the bottom include a large number of `easy' to classify negative values, but the size of the overlap is only slightly increased over the datasets at the top. The data generation method and the code are available at: \href{https://github.com/soccermatics/ABC.git}{https://github.com/soccermatics/ABC.git}.

Across the nine data sets, $A$ is high when the scores of the positive and negative classes are well separated (figure \ref{fig:Distribution}a, d and g, with $A$ between 0.97 and 0.99). But $A$ also increases when easy negatives are added, even though the difficult negatives stay exactly as difficult: going from c through f to i, $A$ rises from 0.68 to 0.85 to 0.96. In data set i, most randomly chosen negatives are easy ones, so a value of $A$ close to one tells us very little about how well the algorithm separates the positives from the difficult negatives. Adding easy cases to a test set can therefore make an algorithm appear to improve when it has not.

\subsection*{Indistinguishability threshold}

In line with the method described in the introduction, we now compare the positive class with the members we have labelled as positive, rather than with the negative class. In the cat labelling problem, we pick an image at random from all images containing cats and denote its score $S_P$. Then, for a given threshold $r$, we pick an image at random from all images labelled as cats, i.e. those with a score above $r$, whether or not they contain a cat, and denote its score $S_R$. We define
\[
B(r) = \prob(S_P>S_R | S_R>r),
\]
In words this is the probability that a random member of the positive class scores higher than a random member of the set labelled positive. When $r$ is large, only a few high-scoring members are labelled positive, so $B(r)$ is close to zero: precision is high, but many positives are missed (false negatives). When $r$ is small, nearly everything is labelled positive, so $B(r)$ is large: few positives are missed, but many negatives are labelled positive (false positives) and precision is low. Between these extremes, $B(r)$ decreases as $r$ increases (see the Appendix).

We now define $r_b$ such that
 \begin{equation}
B(r_b) = \prob(S_P>S_R | S_R>r_b) = 1/2 \label{balanced}
\end{equation}
so that, as far as the classifier is concerned, the images in the positive class are indistinguishable from those which we have labelled positive. If we pick a random member of the images we have classified as positive and a random member of the positives, these two images are equally likely to have the higher score. We can think of $r_b$ as being the threshold value at which the model cannot distinguish between true positives and labelled positives. We therefore refer to $r_b$ as the {\it indistinguishability threshold}. 

The vertical dotted lines in figure \ref{fig:Distribution} show the value of $r_b$ for our nine example score distribution sets. For all nine distributions the lines give a reasonable separation between the scores, lying roughly halfway between the `difficult' negative cases and the positive cases. 

This notion of indistinguishability is related to several existing concepts. It is {\it stochastic equality} \cite{vargha2000critique}, which, unlike identical distributions of scores, allows the members labelled positive to have, for example, a larger spread of scores than the positive class. It is also the property tested by the Mann--Whitney test \cite{mann1947test}, which compares two groups by how often a randomly chosen member of one scores higher than a randomly chosen member of the other. And, in terms of placement values \cite{hanley1997sampling, pepe2004analysis}, $B(r)$ is the mean proportion of the positive class that scores higher than a member labelled positive, so $B(r_b)=1/2$ places the members labelled positive, on average, where the positive class places itself.

A misguided (but still sometimes proposed) alternative to choosing $r=r_b$ is to instead choose $r$ such that
 \begin{equation}
\eta(r) = 1 - \eta(r) \label{standard}
\end{equation}
or, equivalently,
\begin{equation}
\eta(r) = 1/2 \label{standard2}
\end{equation}
where $\eta(r)$ is the probability that a member with score $r$ belongs to the positive class, according to an underlying model which converts scores in to probabilities. For example, when the scores are the log-odds estimated by logistic regression, the model is the logistic function
\[
\eta(r) = \frac{1}{1+\exp(-r)} = 1/2
\]
which gives $r=0$ as the threshold. Looking again at the distribution of scores in figure \ref{fig:Distribution}  we see that the size of the negative class shifts the distribution of the scores generated by the logistic regression to the left. This is undesirable, because it means that the algorithm will make more errors when easy to categorise cases are added to the data set.

The limitations of using equation \ref{standard2} to set a threshold are known, but often they are referred to as being ``caused by an unbalanced data set''. The implication is that too many members in the negative class have somehow misled us in to choosing a poor value of $r$. In our experience, students and even experienced practitioners, when confronted with this problem, either start to prune their data, removing `easy' members of the negative class, or oversample from the positive class in order to compensate. They might also draw an ROC curve and a precision-recall curve and try to justify, in an ad-hoc fashion, choosing a different value of the threshold $r$ on the basis of what they see in these curves.

We would argue that it is not the data set that is imbalanced, but this way of looking at it that is wrong! Specifically, it is the arbitrary choice of $r=0$ that is imbalanced, because it lies a long way from the indistinguishability threshold. We now propose a measure, based on the indistinguishability threshold, that avoids the need for any "rebalancing" of the data set.

\begin{figure}[tp]
\centering
\includegraphics[scale=1]{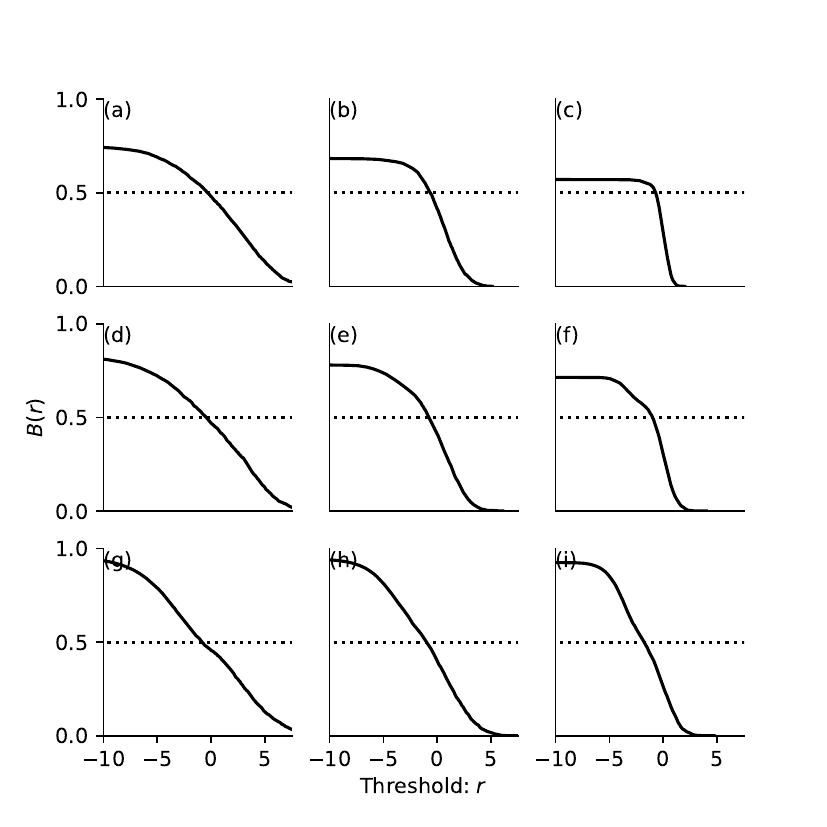}
\caption{Plot of $B(r)$ for our 9 example data sets. The dotted line is at $1/2$ indicates the value at which the model is balanced.}
\label{fig:Bplot}
\end{figure}

\subsection*{Precision at the indistinguishability threshold}

\label{PIT}

How often is our positive labelling correct when reality and labelling are indistinguishable? To answer this, we use the precision, $C(r)$: the proportion of members labelled positive (score greater than $r$) that are in the positive class. There are $P v(r)$ true positives and $N u(r)$ false positives, so
\begin{equation}
C(r) = \frac{P v(r)}{P v(r) + N u(r)}. \label{precisionP}
\end{equation}
When $N$ is much larger than $P$, the true positive rate $v(r)$ has to be much larger than the false positive rate $u(r)$ for precision to be high.

We propose measuring precision specifically with a value of $r=r_b$, i.e. for a threshold whereby data classified as positive cannot be distinguished from data which truly is positive, i.e. we calculate $C(r_b)$, the precision at the indistinguishability threshold (PIT). Evaluating this for the data in figure \ref{fig:Distribution} (a), (d) and (g) we get $C(r_b) \approx  0.87$. This could be considered a reasonably good performance: 87\% of the pictures the algorithm says are cats are in fact cats. For figure \ref{fig:Distribution}  (b), (e) and (h) we get $C(r_b) \approx 0.70$. This is not a great performance, but could be useable in certain applications: 70\% of the pictures the algorithm says are cats are in fact cats. Finally, for the data in figure \ref{fig:Distribution} (c), (f) and (i) we get $C(r_b) \approx  0.53$. This is poor performance, and is probably not very useful in most applications: only 53\% of the pictures the algorithm says are cats are in fact cats.

We can now see why $C(r_b)$ is preferable to $A$ as a measure of performance. When we calculated $A$, the value was highly dependent on the number of `easy' cases. This is not the case with $C(r_b)$, nor, since it also ignores members labelled negative, with the F1-score. By evaluating precision when the algorithm cannot distinguish reality from its own labelling, we obtain a measure of performance that doesn't depend on the number of `easy' cases. The choice of $r_b$ is theoretically motivated (in the previous section) and gives reasonable results. Before looking at how $C(r_b)$ relates to the F1-score, we test it on real data.

\subsection*{Two benchmark data sets}

\label{sec:benchmarks}

On a standard benchmark data set of mammograms, two classifiers with the same area under the curve (difference in $A$ of $-0.003$, 95\% interval $-0.044$ to $0.038$) differ by 13 percentage points in PIT (95\% interval 6 to 21 percentage points), illustrating why $C(r_b)$, rather than $A$, should be used to compare their performance.

In the artificial data sets, adding easy negatives increased $A$ but left $C(r_b)$ almost unchanged. The same holds on real data, and exactly. In the two benchmark data sets below, removing the half of the test negatives with the lowest scores, i.e. the easiest cases, reduces $A$ by between 0.04 and 0.09 but leaves $C(r_b)$ unchanged, since none of these members are labelled positive. The value of $A$ thus depends on how many easy cases happen to be in the test set, while $C(r_b)$ does not. To see how the measure compares classifiers in practice, we used two standard benchmark data sets, both obtained from OpenML \cite{vanschoren2013openml}. The Adult data set \cite{becker1996adult} contains 48,842 individuals from the 1994 US census, of whom 24\% have an annual income above \$50,000 (the positive class). The Mammography data set \cite{woods1993comparative} contains 11,183 regions of digitised mammograms, of which 260 (2.3\%) contain microcalcifications. For each data set, we trained a logistic regression and a gradient boosting classifier on a randomly chosen half of the data (stratified by class) and calculated all measures exactly from the scores on the other half. Confidence intervals were obtained by resampling the test set 1000 times, with both classifiers evaluated on the same resampled test sets. The code is available in the same repository as the code for the artificial data sets. The results are shown in table \ref{tab:benchmarks}.

\begin{table}[tp]
\centering
\footnotesize
\setlength{\tabcolsep}{3.5pt}
\begin{tabular}{lrrcccccc}
\toprule
Model & $P$ & Labelled & $C(r_b)$ & $v(r_b)$ & PRBEP & $F_1(r_b)$ & max $F_1$ & $A$ \\
\midrule
\multicolumn{9}{l}{\it Adult} \\
LR & 5844 & 8305 & 0.58 (0.57--0.59) & 0.83 & 0.68 & 0.69 & 0.69 & 0.91 \\
GB & 5844 & 7532 & 0.64 (0.63--0.65) & 0.83 & 0.72 & 0.72 & 0.73 & 0.93 \\
\multicolumn{9}{l}{\it Mammography} \\
LR & 130 & 260 & 0.35 (0.25--0.45) & 0.71 & 0.56 & 0.47 & 0.60 & 0.92 \\
GB & 130 & 194 & 0.49 (0.36--0.59) & 0.72 & 0.65 & 0.58 & 0.69 & 0.92 \\
\bottomrule
\end{tabular}
\caption{Performance on the test half of two benchmark data sets, for logistic regression (LR) and gradient boosting (GB). $P$ is the number of positives in the test set and `Labelled' the number labelled positive at the indistinguishability threshold $r_b$. $C(r_b)$ is given with a 95\% bootstrap interval. PRBEP is the precision (equal to the recall) at the precision-recall break-even point.}
\label{tab:benchmarks}
\end{table}

The measure can be reported in the same plain language as in the rest of this article. For gradient boosting on the Mammography data, for example: {\it when the algorithm is tuned so that the regions it flags cannot be distinguished from the regions that really contain microcalcifications, 49\% of the flagged regions contain microcalcifications} (95\% confidence interval 36--59\%, compared with 2.3\% if regions were flagged at random), and 72\% of the regions containing microcalcifications are flagged.

\subsection*{Tuning false positive and true positive rates}

There are two further questions we might ask. 
\begin{enumerate}
\item How often is a member of the positive class detected by the test? This is the true positive rate, $v(r)$.
\item How often is a member of the negative class labelled positive by the test? This is the false positive rate, $u(r)$.
\end{enumerate}
In medical testing, the true positive rate is called the {\it sensitivity} and one minus the false positive rate is called the {\it specificity} \cite{altman1994diagnostic}. Choosing a threshold $r$ amounts to choosing an {\it operating point} on the ROC curve, trading one rate against the other \cite{fawcett2006introduction}.
Typically, one of these rates will be more important than another. For example, we might, in a medical screening task, accept a higher false positive rate in order to make sure all those who are positive for a condition are identified. In this case we set $r$ to be a smaller value so as to get a high true positive rate, but also (possibly) a high false positive rate.  In a targeted advertising task, where people who might be interested in a product should be identified, we might concentrate on having a lower false positive rate and accept a lower true positive rate, identifying only individuals who are very likely to be interested in the advertised product. In this case we set $r$ to a larger value. The literature offers several ways of making this choice. If the costs of a false positive and of a false negative can be stated, the threshold can be chosen to minimise the expected cost \cite{elkan2001foundations}. If, instead, there is an upper limit on how often negatives may be labelled positive, the threshold can be chosen to make the true positive rate as large as possible while keeping the false positive rate below that limit. This is known as {\it Neyman--Pearson classification} \cite{neyman1933ix, scott2005neyman, tong2018neyman}. Youden's index, $v(r)-u(r)$, is often maximised when neither rate is given priority \cite{youden1950index}. In general, however, there is no escaping from the fact that the measure of performance, and thus the choice of $r$, is informed by the application area.

\begin{figure}[tp]
\centering
\includegraphics[scale=1]{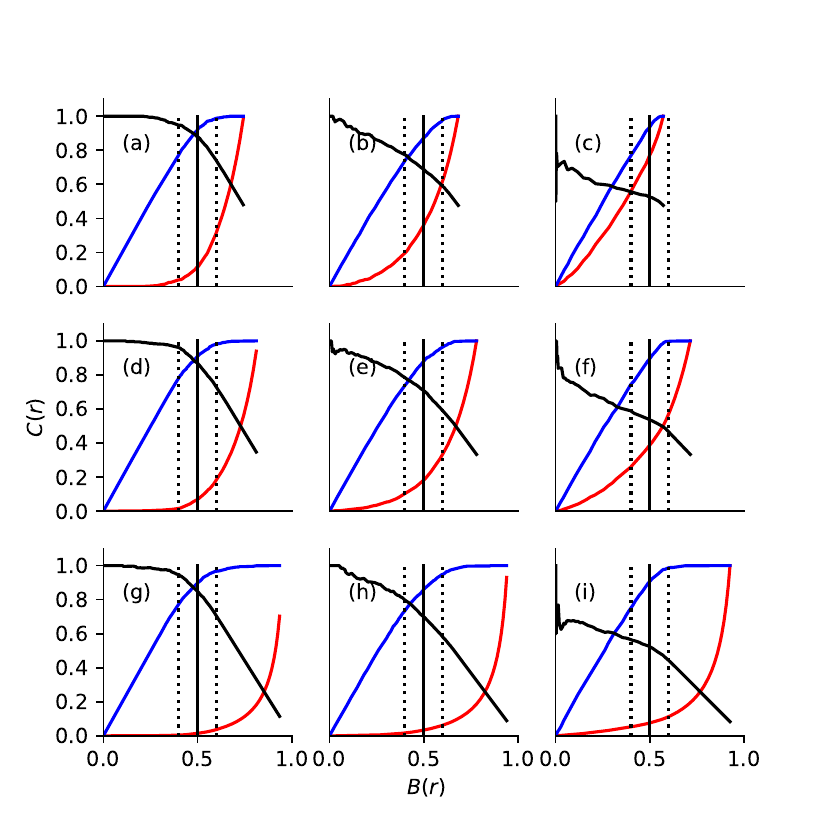}
\caption{Plot of $B(r)$ against the precision $C(r)$ (black line), true positive rate $v(r)$ (blue) and false positive rate $u(r)$ (red) for our 9 example data sets. These curves are parameterised by $r$. }
\label{fig:balance}
\end{figure}

An alternative here is to set an acceptable range for having a threshold which does not preserve the property that reality is indistinguishable from labelling. A rule of thumb is 40/60. If, for example, we want to ensure a high level of true positives, we find $r_{60}$ by solving $B(r_{60})=0.6$, so that the probability a member classified as positive has a lower score than a truly positive member is 60\%. This is shown as the dotted lines in figure \ref{fig:balance} to the right of the solid line (at which $B(r_{b})=1/2$). Similarly, we can find $r_{40}$ by solving $B(r_{40})=0.4$, which is given to the left. Note that, while $B(r_{60})>B(r_b)>B(r_{40})$, the thresholds have the relationship $r_{60}<r_b<r_{40}$. A lower threshold leads to more true positives and more false positives.

Figure \ref{fig:balance} shows $C(r)$ (black line) plotted against $r$ for our nine test data sets. Again, the way we built $B(r)$ allows us to explain in words what we are doing here. We can, for example, write about figure \ref{fig:balance}(e) that, "accepting a 60/40 imbalance in favour of positives our algorithm's predictions in favour of the positive class gave us a precision of $C(r_{60})=0.60$, compared to $C(r_b)=0.71$ at 50/50 balanced predictions."

\subsection*{Putting the F1-score into words}

\label{sec:others}

Precision and recall pull in opposite directions. Lowering the threshold finds more members of the positive class, increasing recall, $v(r)$, but also labels more negatives as positive, reducing precision, $C(r)$. Reporting only one of them can therefore be misleading: labelling every member positive gives perfect recall, and labelling only the single highest-scoring member can give perfect precision. The F1-score combines the two into a single number. It is their harmonic mean,
\begin{equation}
F_1(r) =  \frac{2 C(r) v(r)}{C(r) + v(r)} .  \label{eq:F1}
\end{equation}
It is high only when both precision and recall are high. This is why the F1-score is widely used, particularly in information retrieval and natural language processing \cite{chinchor1993muc, manning2008introduction, sasaki2007truth}, and why the threshold which maximises it is a common choice \cite{lipton2014optimal}. But, as we noted in the Introduction, it has no simple interpretation in words \cite{hand2018note}. In this section, we use $B(r)$ to give it one.

\begin{figure}[tp]
\centering
\includegraphics[scale=1]{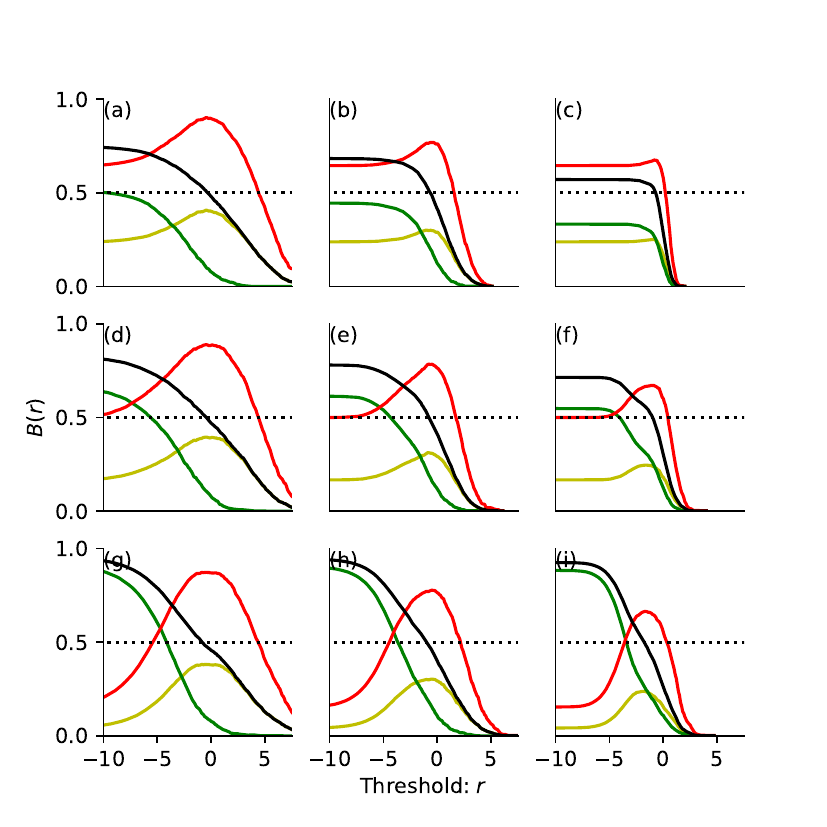}
\caption{Plot of $B(r)$ for our 9 example data sets. The black line is $B(r)$. The dotted line at $1/2$ indicates the value at which the model is balanced. The yellow line is the component $B_+(r)$ and the green line is $B_-(r)$ (equation \ref{eq:Bexact}).  The red line is the $F_1(r)$ score (equation \ref{eq:F1}).}
\label{fig:Bplot2}
\end{figure}

Like the F1-score, $B(r)$ combines precision and recall, as we can see by looking more closely at its components. Let
\begin{equation}
\alpha(r) = \prob(S_P > S_N \mid S_P > r, S_N > r)  \label{eq:alpha}
\end{equation}
be the probability that a randomly chosen true positive scores higher than a randomly chosen false positive. In other words, $\alpha(r)$ is the area under the curve computed using only those members labelled positive, so that $\alpha(-\infty)=A$. Now, the member labelled positive in the comparison that defines $B(r)$ is either a true positive, with probability $C(r)$, or a false positive, with probability $1-C(r)$. In both cases, the member of the positive class can only score higher if it also scores above $r$, which happens with probability $v(r)$. If the labelled member is a true positive, both are then members of the positive class scoring above $r$, and each is equally likely to have the higher score. If it is a false positive, the member of the positive class has the higher score with probability $\alpha(r)$. Combining the two cases gives
\begin{equation}
B(r) = v(r) \left( \frac{C(r)}{2} + \alpha(r) \left(1 - C(r)\right) \right) = B_+(r) + B_-(r),  \label{eq:Bexact}
\end{equation}
where $B_+(r) = v(r) C(r)/2$ is the contribution from true positives and $B_-(r) = v(r) \alpha(r) (1 - C(r))$ the contribution from false positives. (In the Appendix we derive the same expression by integrating over the score distributions.) Thus, like the F1-score, $B(r)$ is built from recall and precision: it is recall multiplied by a weighted average of $1/2$ and $\alpha(r)$, with weights given by the precision. Figure \ref{fig:Bplot2} shows $B(r)$, its two components and the F1-score for our nine data sets.

The first of these terms is closely related to the F1-score. Figure \ref{fig:Bplot2} shows that $B_+(r)$ (yellow line) reaches its maximum close to the threshold at which the F1-score (red line) is maximised. To see why, we differentiate both with respect to $r$ (see Appendix). The two maxima satisfy the same condition, except that the F1-score appears in one where precision appears in the other \cite{lipton2014optimal}. Since the F1-score lies between precision and recall, the two maxima therefore coincide at a threshold at which precision equals recall, and lie close to each other when precision and recall are similar there. Moreover, the F1-score is flat near its maximum, so the F1-score at the threshold which maximises $B_+(r)$ is close to its maximal value even when the two thresholds differ. On the benchmark data sets (table \ref{tab:benchmarks}), it is within 0.005 of the maximal F1-score in all four cases.

This allows us to put the F1-score into words. The product $v(r) C(r) = 2 B_+(r)$ is the probability that, if we pick one image that really contains a cat and one image labelled as a cat, the image containing the cat is labelled as a cat and the labelled image does contain a cat: both picks are hits. Maximising the F1-score approximately maximises this probability. In words, then, {\it the F1-score chooses, approximately, the threshold at which a randomly chosen real cat and a randomly chosen labelled image are most likely both to be hits}. This, we suggest, is part of why the F1-score works as well as it does in practice. Our interpretation differs from that of Hand, Christen and Kirielle \cite{hand2021f}, who transform the F1-score into $F^* = F_1/(2-F_1)$, the proportion of members that are either labelled positive or in the positive class which are both. Their interpretation applies to the value of the F1-score at any threshold. Ours applies to the threshold at which the F1-score is maximised, and explains why that threshold is a reasonable choice.

What this interpretation leaves out is how the algorithm ranks its false positives. The indistinguishability threshold, which includes them, is chosen in a different way. It is not the maximum of anything, since $B(r)$ decreases as $r$ increases, but the threshold at which the whole of $B(r)$ equals one half. The second term, $B_-(r) = v(r) \alpha(r) (1 - C(r))$, accounts for how the algorithm ranks its false positives relative to the positive class, and has no counterpart in the F1-score. In this limited sense the F1-score is a surrogate for PIT. It looks for a threshold using only precision and recall, which make up the first part of $B(r)$, and summarises performance as a mean of the two, rather than answering a question that can be stated in words. Including $B_-(r)$ changes the question from \lq at which threshold are both picks most likely to be hits?\rq\ to \lq at which threshold can the algorithm no longer distinguish labelled positives from real ones?\rq. The consequences of this change for precision, recall and the F1-score can be derived exactly.

Setting $B(r_b)=1/2$ in equation \ref{eq:Bexact} and solving for the recall gives
\begin{equation}
v(r_b) = \frac{1}{C + 2\alpha(1-C)} ,  \label{eq:vrb}
\end{equation}
where $C=C(r_b)$ and $\alpha=\alpha(r_b)$. So, at the indistinguishability threshold, recall is determined by precision and $\alpha$ alone. This has two consequences. First, since recall cannot exceed one, $\alpha \geq 1/2$: true positives outscore false positives at least as often as would be expected by chance. Second, since $\alpha \leq 1$, the denominator is at most $2-C$, and so $v(r_b) \geq 1/(2-C) > C$ whenever $C<1$. Recall is therefore always larger than precision at the indistinguishability threshold. Equivalently, more members are labelled positive than there are members of the positive class, and $r_b$ always lies below the precision-recall break-even point, the threshold at which precision equals recall \cite{manning2008introduction}. Substituting equation \ref{eq:vrb} into equation \ref{eq:F1} gives the F1-score at $r_b$,
\begin{equation}
F_1(r_b) = \frac{2C}{1 + C^2 + 2\alpha C (1-C)} .  \label{eq:F1atrb}
\end{equation}

Equation \ref{eq:vrb} also shows how far below the break-even point the indistinguishability threshold lies, and hence where the F1-score and PIT diverge. The ratio of recall to precision at $r_b$ is $1/(C(C + 2\alpha(1-C)))$, which grows as precision falls. The threshold which maximises the F1-score, in contrast, tends to lie nearer the break-even point, where it coincides with the maximum of $B_+(r)$. In the examples in figure \ref{fig:Bplot2} the two thresholds are close, but this is not true in general: they diverge most when precision is low.

Equation \ref{eq:vrb} also shows how PIT differs from the precision at the break-even point. PRBEP is the precision when exactly as many members are labelled positive as there are positives, and so does not depend on how the false positives are ranked relative to the true positives. PIT does, through $\alpha$. For a given precision, if the false positives score well below the true positives ($\alpha$ close to one), recall at $r_b$ is close to $1/(2-C)$, and $r_b$ lies relatively near the break-even point. If the false positives are mixed in among the true positives ($\alpha$ close to one half), recall at $r_b$ approaches one: nearly all positives must be labelled before the labelled set becomes indistinguishable from the positive class, and PIT is correspondingly lower. PIT thus combines precision with how well the algorithm separates the members it correctly labels positive from those it wrongly labels positive, which PRBEP does not take into account.

The benchmark results in table \ref{tab:benchmarks} confirm these properties. In all four cases more members are labelled positive at $r_b$ than there are members of the positive class, so that recall is larger than precision and PIT is between 0.08 and 0.21 below PRBEP. PIT also separates the two classifiers by more than PRBEP does: by 0.06 rather than 0.04 on the Adult data, and by 0.14 rather than 0.09 on the Mammography data. Recall at $r_b$ agreed with equation \ref{eq:vrb} to within $0.004$ in all four cases. On the Adult data, where PIT is about 0.6, the F1-score at $r_b$ is within 0.005 of its maximum. On the Mammography data, where PIT is 0.35--0.49, it is 0.11--0.13 lower, and for logistic regression twice as many regions are labelled positive at $r_b$ as there are positives. Here the maximal F1-score ranks the two classifiers in the same order as PIT, whereas the AUC cannot tell them apart, but the two measures choose very different thresholds.

\section*{Discussion}

Every quantity in this article answers a question that can be stated in words, and every answer is either a threshold or a proportion that can be reported as a percentage (table \ref{tab:words}). Even the exact results can be stated this way: at the indistinguishability threshold, the algorithm always labels more images as cats than there are cats, so the proportion of cats it finds is always larger than the proportion of labelled images that are cats. The F1-score, by contrast, does not by itself answer any question that a user of the algorithm would naturally ask \cite{hand2018note, hand2021f}, but our approach allows us to put it into words.

\begin{table}[tp]
\centering
\small
\begin{tabular}{p{2.3cm}p{9.4cm}}
\toprule
Measure & Question it answers \\
\midrule
$A$ (AUC) & If the algorithm is shown one image that contains a cat and one that does not, how often does it give the higher score to the image containing the cat? \\[3pt]
$B(r)$ & If we pick one image labelled as a cat and one image that really contains a cat, how often does the real cat have the higher score? \\[3pt]
$r_b$ & At what threshold is an image labelled as a cat equally likely as an image of a real cat to have the higher score? \\[3pt]
$C(r_b)$ (PIT) & When the algorithm is tuned so that it can no longer distinguish the images it labels as cats from images that really contain cats, how often does an image labelled as a cat actually contain one? \\[3pt]
$v(r_b)$ & At that threshold, what proportion of the images containing cats are labelled as cats? \\[3pt]
$\alpha(r_b)$ & Among the images labelled as cats, how often does an image containing a cat have a higher score than an image that does not? \\[3pt]
$C(r_{60})$ & If we accept that a real cat scores higher than an image labelled as a cat 60\% of the time, how often does an image labelled as a cat actually contain one? \\[3pt]
$r$ maximising $F_1$ & (Approximately) At what threshold is it most likely that a randomly chosen real cat is labelled as a cat and a randomly chosen image labelled as a cat contains one? \\
\bottomrule
\end{tabular}
\caption{The measures used in this article and the questions they answer. Each answer is either a threshold or a proportion that can be reported as a percentage.}
\label{tab:words}
\end{table}

Our approach starts from a statement of what we want to know, a threshold at which the images labelled as cats cannot be distinguished from those that really contain cats, and then writes this requirement out as $B(r_b)=1/2$. We then see that it splits into two parts. The first, $B_+(r)$, depends only on precision and recall, and maximising the F1-score approximately maximises it. This puts the F1-score into words: its threshold approximately maximises the probability that a randomly chosen real cat is labelled as a cat and a randomly chosen labelled image is a cat, which, we suggest, is part of why it works well in practice. The second, $B_-(r)$, depends on how the false positives are ranked and has no counterpart in the F1-score. Including it changes the threshold from a maximum to the point at which labelling and reality are indistinguishable. This point always lies below the precision-recall break-even point, and furthest below it when precision is low, as in the Mammography data. Where the F1-score is currently reported, we suggest reporting PIT instead, since it states in words what the F1-score leaves implicit.

The break-even point offers a different comparison. Like PIT, it matches the set labelled positive to the positive class, and PRBEP has a simple reading in words: if the algorithm labels as many images as cats as there are real cats, what proportion of the labelled images contain one? This reading is, if anything, simpler than ours. PIT differs in matching the scores of the two sets rather than their sizes. As a result, it always labels more members positive than there are positives, so that fewer positives are missed, and it takes into account how well the algorithm separates its correct labels from its false alarms. Matching sizes is a convention; matching scores means that the algorithm cannot tell its own labelling from reality. We argue that the latter is the more natural basis for a threshold, but the choice between the two is ultimately one of which question a user wants answered.

Being able to state a measure in words matters because those who decide whether to use an algorithm, such as clinicians, managers and regulators, are rarely machine learning specialists. For the Mammography data, we can tell a radiologist that, at the indistinguishability threshold, about half of the regions flagged by gradient boosting contain microcalcifications, so each correct detection comes with roughly one false alarm, and that 72\% of the regions with microcalcifications are found. These statements translate directly into the workload the tool creates and the cases it misses. The AUC of 0.92, by contrast, sounds excellent, but answers a question a radiologist never faces and hides a large difference between the two classifiers. Statements in words also help when the costs of the two types of error, which in medicine in particular should be weighed against each other \cite{neyman1933ix, scott2007performance, obuchowski2018receiver, chu1999introduction, care2018new, elkan2001foundations}, have not yet been specified. We can say, for example, that accepting that a real tumour scores higher than an image labelled as a tumour 60\% of the time, rather than 50\%, finds more of the tumours but reduces the proportion of labelled images that are tumours from 71\% to 60\% (figure \ref{fig:balance}e).

Like the threshold that maximises the F1-score, the indistinguishability threshold depends only on the ranking of the scores, so the scores need not be calibrated probabilities, and it is not affected by adding members that are easy to classify. PIT also respects ROC dominance: if one classifier's ROC curve lies above another's, and either curve is concave, the first has the higher PIT (see Appendix). For ROC curves that are not concave this can fail, unlike for the maximal F1-score and PRBEP, but only where precision rises as more members are labelled positive. The measure also has limitations. Like any precision, PIT depends on the relative numbers of positive members and of negative members that are difficult to classify, so it should be compared across data sets with care; but this is also what tells users how many flagged cases will be correct in their own population. This property could potentially make the methods described here useful beyond evaluation, for example in boosting \cite{schapire2003boosting}, by repeatedly fitting new models to the members scoring above $r_b$.

Many machine learning algorithms are developed before the trade-off between error types in an application has been specified. In such situations, PIT combines the advantages of AUC, that it can be explained in words, and of the F1-score, that it is not inflated by easy cases. We would therefore encourage researchers to report PIT, together with the recall at the same threshold, and to state both in words: {\it when the algorithm is tuned so that it can no longer distinguish the cases it flags from real cases, X\% of the flagged cases are real, and Y\% of the real cases are flagged.} Reported in this way, the performance of an algorithm can be understood, questioned and discussed by the people who will use it.

\section*{Acknowledgements}

Thanks to Daniel Gedon and Andre Teixeira for feedback. 

\section*{Appendix: Properties and implementation of $B(r)$}
\label{app:B}

In this appendix we show how $B(r)$ can be calculated. As with $A$ the pairwise nature of the test means it is straightforward to implement a randomised version of $B(r)$. For each value of $r$ we sample, with or without replacement, $n$ members of the positive class. Then we sample $n$ members from the set labelled positive when the threshold is greater than $r$. Pairwise comparison of the scores of these two sets gives $B(r)$ . This method becomes less reliable for large values of $r$, since the size of the positive labelled class becomes small and all the comparisons will be made to a small number of positively labelled images. 

Another way to calculate $B(r)$, following the AUC approach to estimating $A$, is by integration over the true positive rate, $v(r)$, and the false positive rate, $u(r)$. We now derive the appropriate integral. First, let $S_R$ be the score of a randomly chosen member of the data set and $S_P$ the score of a randomly chosen member of the positive class. The probability that $S_R$ is less than $S_P$, given that $S_R$ is greater than $r$, is then written as $\prob(S_R<S_P | S_R>r)$. It is this probability we call $B(r)$.

Our approach to finding an expression for $B(r)$ is to look separately at the case where $S_R$ is from a negative member and a positive member. We can use the law of total probability to give 
\begin{eqnarray}
B(r) &= &\prob(S_R < S_P | S_R>r) \nonumber \\
& = & \prob(S_R < S_P | S_{R} > r , R = + ) \cdot \prob( R = + | S_R > r )  \nonumber \\
&&  + \prob(S_R < S_P | S_R>r, R= - ) \cdot \prob( R = - | S_R>r ) \label{eqn:totprobstep} 
\end{eqnarray}
where $+$ and $-$ are respectively the events that the chosen individuals are in the positive and negative classes. If an individual is chosen at random then 
\begin{equation}
\prob( R = + | S_R>r ) =   \frac{P v(r)}{P v(r) + N u(r)}  \mbox{ and }  \prob( R = - | S_R>r ) =   \frac{N u(r)}{P v(r) + N u(r)} \label{eqn:probbe}
\end{equation}
The first of these is the precision, $C(r)$, given in equation \ref{precisionP}. Furthermore,
\begin{eqnarray}
\prob(S_R < S_P | S_{R} > r ,  R = +   ) & = & \frac{\prob(S_R < S_P \cap S_{R} > r ,  R = +  )}{\prob(S_{R} > r ,  R = +  )} \nonumber \\
& = & \frac{\int_{t=r}^\infty \int_{s=t}^\infty f_P(t) f_P(s) ds dt}{\int_{t=r}^\infty f_P(t) dt} \nonumber \\
& = & \frac{\int_{t=r}^\infty f_P(t) v(t) dt}{v(r)} \label{eqn:posstep}
\end{eqnarray}
Since the derivative of $v(r)$ is equal to $-f_P(r)$ we can change variable to $v$ in the integral in the numerator to get
 \begin{eqnarray}
\int_{t=r}^\infty f_P(t) v(t) dt =  \int_{t=r}^\infty v(t) \left( -v'(t) \right) dt = \int_{0}^{v(r)} v \, dv = \frac{v(r)^2}{2}
\end{eqnarray}
and thus simplify to get
\begin{equation}
\prob(S_R < S_P | S_{R} > r ,  R = +   ) = \frac{v(r)}{2} \label{eqn:posstepfinal}
\end{equation}
We can follow similar steps for the case that $R$ is negative to get,
\begin{equation}
\prob(S_R < S_P | S_{R} > r ,  R = -   ) = \frac{\int_{t=r}^\infty f_N(t) v(t) dt}{u(r)}  \label{eqn:negstep}
\end{equation}
Substituting \ref{eqn:probbe}, \ref{eqn:posstepfinal} and \ref{eqn:negstep} in to equation \ref{eqn:totprobstep} gives
\begin{equation}
B(r) = \frac{P v(r)^2/2 + N \int_{t=r}^\infty f_N(t) v(t) dt }{P v(r) + N u(r)} .
 \label{eqn:Bdef}
\end{equation}
The numerator in equation \ref{eqn:negstep} is $\prob(S_P > S_N, S_N > r)$, since $S_P>S_N>r$ implies that $S_P>r$. Dividing by $\prob(S_P>r)\prob(S_N>r) = v(r) u(r)$ gives the integral form of equation \ref{eq:alpha},
\begin{equation}
\alpha(r) = \frac{\int_{t=r}^\infty f_N(t) v(t) dt}{u(r) v(r)} ,
\end{equation}
so that equation \ref{eqn:negstep} equals $\alpha(r) v(r)$ and equation \ref{eqn:Bdef} is the same as equation \ref{eq:Bexact}. Equation \ref{eqn:Bdef} is plotted as the black line in figure \ref{fig:Bplot}. As $r \rightarrow -\infty$, $v(r)$ and $u(r)$ tend to one and $\int_{t=r}^\infty f_N(t) v(t) dt$ tends to $A$ (equation \ref{eq:SPgreaterSN2}), so that
\[
B(-\infty) = \frac{P/2 + N A}{N + P} 
\]
while $B(r) \leq v(r) \rightarrow 0$ as $r \rightarrow \infty$. 

In between, $B(r)$ always decreases as $r$ increases. To see this, write $D(r) = P v(r) + N u(r)$ for the denominator of equation \ref{eqn:Bdef}. The derivative of the numerator is $P v(r) v'(r) - N f_N(r) v(r) = v(r) D'(r)$, so that
\[
B'(r) = \frac{D'(r)}{D(r)} \left( v(r) - B(r) \right).
\]
Since $D'(r) = -(P f_P(r) + N f_N(r)) \leq 0$ and, from equation \ref{eq:Bexact}, $B(r) \leq v(r) \left(1 - C(r)/2\right) < v(r)$, we have $B'(r) \leq 0$. Thus $B(r)$ decreases from $B(-\infty)$ to zero, and a threshold $r_b$ with $B(r_b)=1/2$ exists if and only if $B(-\infty) \geq 1/2$, which is equivalent to $A \geq 1/2$. In other words, the indistinguishability threshold exists whenever the algorithm does at least as well as random guessing, and it is unique wherever the densities of scores are positive. It is identified as the point in figure \ref{fig:Bplot} where $B(r)$ crosses the dotted line at $1/2$, and is plotted as the vertical dotted line in figure \ref{fig:Distribution}.

\subsection*{The maxima of the F1-score and of $B_+(r)$}

Write $D(r) = P v(r) + N u(r)$, as above, and
\[
\eta(r) = \frac{P f_P(r)}{P f_P(r) + N f_N(r)}
\]
for the probability that a member with score exactly $r$ is in the positive class. Since $v'(r) = -f_P(r)$ and $D'(r) = -(P f_P(r) + N f_N(r))$, we have $P v'(r) = \eta(r) D'(r)$.

The F1-score can be written as $F_1(r) = 2 P v(r) / (D(r) + P)$. Its derivative is zero when $v'(r)(D(r)+P) = v(r) D'(r)$, i.e. when $\eta(r) D'(r) (D(r)+P) = P v(r) D'(r)$. Dividing by $D'(r)$ gives
\begin{equation}
\eta(r) = \frac{P v(r)}{D(r) + P} = \frac{F_1(r)}{2} , \label{eq:F1max}
\end{equation}
which is the result of Lipton et al.\ \cite{lipton2014optimal}: the F1-score is maximised at the threshold at which the probability of being positive equals half of the maximal F1-score.

Similarly, $B_+(r) = P v(r)^2 / (2 D(r))$, and its derivative is zero when $2 v'(r) D(r) = v(r) D'(r)$, i.e. when $2 \eta(r) D'(r) D(r) = P v(r) D'(r)$, which gives
\begin{equation}
\eta(r) = \frac{P v(r)}{2 D(r)} = \frac{C(r)}{2} . \label{eq:Bpmax}
\end{equation}
Equations \ref{eq:F1max} and \ref{eq:Bpmax} differ only in that the F1-score in the first is replaced by the precision in the second. Since the F1-score is the harmonic mean of precision and recall, $F_1(r) = C(r)$ if and only if $C(r) = v(r)$. A threshold at which $B_+(r)$ is maximised is therefore also a stationary point of the F1-score if and only if precision equals recall there.

\subsection*{PIT and ROC dominance}

Using the change of variable $x = u(t)$ from the section {\it Area under the curve}, the integral in equation \ref{eqn:Bdef} is a partial area under the ROC curve,
\[
\int_{t=r}^\infty f_N(t) v(t) dt = \int_{x=0}^{u} \mathrm{ROC}(x) dx = \mathrm{pAUC}(u),
\]
where $u = u(r)$. Writing $v = \mathrm{ROC}(u)$, $B$ can thus be written as a function of the false positive rate,
\[
B(u) = \frac{P v^2/2 + N\, \mathrm{pAUC}(u)}{P v + N u},
\]
which increases with $u$, since $B(r)$ decreases with $r$. Similarly, $C = P v / (P v + N u)$ depends only on the ratio $v/u$, and increases with it.

Now suppose that the ROC curve of classifier 1 dominates that of classifier 2, i.e. $\mathrm{ROC}_1(x) \geq \mathrm{ROC}_2(x)$ for all $x$. At a fixed $u$, $B$ increases with $\mathrm{pAUC}(u)$, and also with $v$, since
\[
\frac{\partial B}{\partial v} = \frac{P\left(P v^2/2 + N u v - N\, \mathrm{pAUC}(u)\right)}{(P v + N u)^2} > 0 ,
\]
because $\mathrm{pAUC}(u) \leq u v$ for a non-decreasing ROC curve. Both are at least as large for classifier 1, so $B_1(u) \geq B_2(u)$ for all $u$. Since $B$ increases with $u$, classifier 1 reaches $B = 1/2$ first: $u_1 \leq u_2$, where $u_1$ and $u_2$ are the false positive rates at the two indistinguishability thresholds. Then
\[
\frac{\mathrm{ROC}_1(u_1)}{u_1} \geq \frac{\mathrm{ROC}_2(u_1)}{u_1} \geq \frac{\mathrm{ROC}_2(u_2)}{u_2},
\]
where the first inequality follows from dominance, and the second holds whenever $\mathrm{ROC}_2(x)/x$ is non-increasing, which is the case when $\mathrm{ROC}_2$ is concave. (The same conclusion follows if $\mathrm{ROC}_1$ is concave, by applying dominance at $u_2$ instead.) Since precision increases with $v/u$, PIT is at least as high for classifier 1.

A concave ROC curve corresponds to precision that never rises as the threshold is lowered, as is the case when higher scores are always more likely to belong to the positive class. If neither curve is concave, the second inequality can fail, and the dominated classifier can, in exceptional cases, have the higher PIT. In contrast, the maximal F1-score, $\max_u 2 P v / (P v + N u + P)$, and PRBEP, the value of $v$ where the ROC curve crosses the line $P v + N u = P$, can only increase when the ROC curve is raised, and so always respect ROC dominance.

\bibliographystyle{ieeetr}

\bibliography{ref}

\end{document}